\documentclass[journal]{IEEEtran}

\ifCLASSINFOpdf
\else
\fi
\usepackage{amssymb}
\usepackage{amsfonts}
\usepackage{color}
\usepackage{graphicx}
\usepackage{subcaption} 
\usepackage{url}
\usepackage{listings}

\usepackage{xcolor}
\usepackage{amsmath}

\definecolor{codegreen}{rgb}{0,0.6,0}
\definecolor{codegray}{rgb}{0.5,0.5,0.5}
\definecolor{codepurple}{rgb}{0.58,0,0.82}
\definecolor{backcolour}{rgb}{0.95,0.95,0.92}

\lstdefinestyle{mystyle}{
    backgroundcolor=\color{backcolour},   
    commentstyle=\color{codegreen},
    keywordstyle=\color{magenta},
    numberstyle=\tiny\color{codegray},
    stringstyle=\color{codepurple},
    basicstyle=\ttfamily\footnotesize,
    breakatwhitespace=false,         
    breaklines=true,                 
    captionpos=b,                    
    keepspaces=true,                 
    numbers=left,                    
    numbersep=5pt,                  
    showspaces=false,                
    showstringspaces=false,
    showtabs=false,                  
    tabsize=2
}

\begin{document}
%
\title{GEX: Democratizing Dexterity with Fully-Actuated Dexterous Hand and Exoskeleton Glove}
%
%
%

\author{Yunlong Dong$^{*\dagger}$,
        Xing Liu$^*$,
        Jun Wan$^*$,
        Zelin Deng$^*$
\thanks{$^*$ Equal contribution.}
\thanks{Yunlong Dong is an individual researcher.}
\thanks{Xing Liu and Jun Wan are with Huazhong University of Science and Technology.}
\thanks{$^\dagger$Corresponding author, \texttt{yunlongdong@outlook.com}}}

%
%

\markboth{}%
{}
%



\maketitle

\begin{abstract}
This paper introduces GEX, an innovative low-cost dexterous manipulation system that combines the GX11 tri-finger anthropomorphic hand (11 DoF) with the EX12 tri-finger exoskeleton glove (12 DoF), forming a closed-loop teleoperation framework through kinematic retargeting for high-fidelity control. 
Both components employ modular 3D-printed finger designs, achieving ultra-low manufacturing costs while maintaining full actuation capabilities. 
Departing from conventional tendon-driven or underactuated approaches, our electromechanical system integrates independent joint motors across all 23 DoF, ensuring complete state observability and accurate kinematic modeling. 
This full-actuation architecture enables precise bidirectional kinematic calculations, substantially enhancing kinematic retargeting fidelity between the exoskeleton and robotic hand. 
The proposed system bridges the cost-performance gap in dexterous manipulation research, providing an accessible platform for acquiring high-quality demonstration data to advance embodied AI and dexterous robotic skill transfer learning.
The code is published in\footnote{\url{https://github.com/orgs/Democratizing-Dexterous/repositories}}.
\end{abstract}


%
\IEEEpeerreviewmaketitle

\section{Introduction}
Hand dexterity is fundamental to human cognition, enabling active manipulation, tool use, and the way we learn from our environment. 
Replicating this level of dexterity in robotic systems remains a significant challenge~\cite{akkaya2019solving}. 
Although machine learning has driven notable advances in robotic locomotion across varied terrains, real‐world manipulation has largely been limited to simple, single‐degree‐of‐freedom parallel‐jaw grippers~\cite{chi2024universal}. 
Truly dexterous in‐hand manipulation has been investigated primarily in simulation, with only a handful of demonstrations on physical hardware~\cite{shadowhand_inhand}.

A major barrier to widespread research in dexterous manipulation is the lack of affordable, robust robotic hands~\cite{leaphand}. 
High‐performance tendon‐driven designs such as the Shadow Hand achieve human‐like capabilities but carry price tags well over six figures and require substantial maintenance due to their mechanical complexity. 
For many researchers, direct‐drive hands offer a more practical alternative. 
The Allegro Hand, in particular, has seen broad adoption: however, it remains expensive—retailing for over \$16,000.
Consequently, only a small number of laboratories can access hardware capable of complex dexterous tasks. 
In contrast, the ubiquity of affordable two‐finger grippers and legged locomotion platforms has accelerated progress by enabling easy replication and iterative improvement. 
To democratize research in learned dexterous manipulation, a robotic hand platform must be durable, repeatable, low cost, versatile, and anthropomorphic, thus lowering the barrier for broader community engagement.

We introduce the GEX Series, a general‐purpose, low‐cost robotic dexterous hand with 11 DoF and a complementary force‐feedback exoskeleton glove, both constructed from off‐the‐shelf or 3D‐printed parts as robust platforms for robot learning. 
Each device can be assembled in under four hours for \$600—making the hand 1/20 the cost of the Allegro Hand and 1/125 that of the ShadowHand—while the glove likewise takes only hours to build for under \$600.

While robustness and low cost are important, they should not come at the expense of dexterity and anthropomorphism. We believe that a truly versatile system must combine a nimble, human‐like hand with an equally responsive teleoperation interface. 
The human hand can adeptly manipulate a vast array of tools and objects; accordingly, the GX11 hand is designed with an anthropomorphic three‐finger structure that maximizes flexibility while closely mirroring human kinematics. 
Experimental results confirm that our three‐finger dexterous hand achieves manipulation capabilities.

Efficiently collecting high-quality demonstrations remains a persistent challenge in robotic learning. Teleoperation is a common solution, but the high DoF and intrinsic complexity of dexterous hands impose stringent demands on control precision and motion-capture fidelity~\cite{qin2023anyteleop}. 
To address this, our complementary EX12 exoskeleton glove to deliver intuitive, high-fidelity capture of human finger motions, enabling ultra-precise remote operation of the GX11 hand and seamless transfer of demonstrations to the robot platform.
All mechanical designs, embedded firmware, assembly instructions, URDF models, and dexterous retargeting algorithm are released open‐source.


\section{GEX Hardware}
\subsection{Dexterous Hand}
\begin{figure}
    \centering
    \includegraphics[width=0.6\linewidth]{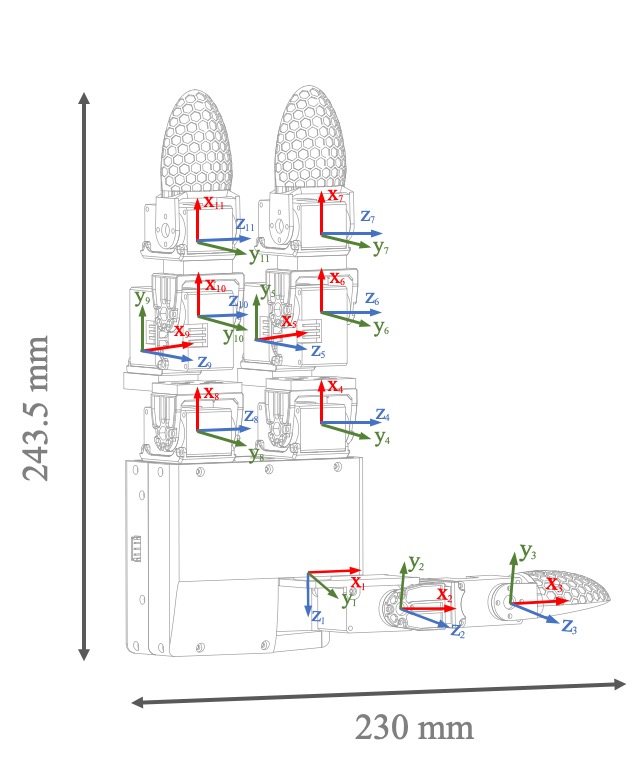}
    \caption{The joint configuration of the GX11 hand.}
    \label{fig:gx11_config}
\end{figure}

\begin{figure}
    \centering
    \includegraphics[width=0.6\linewidth]{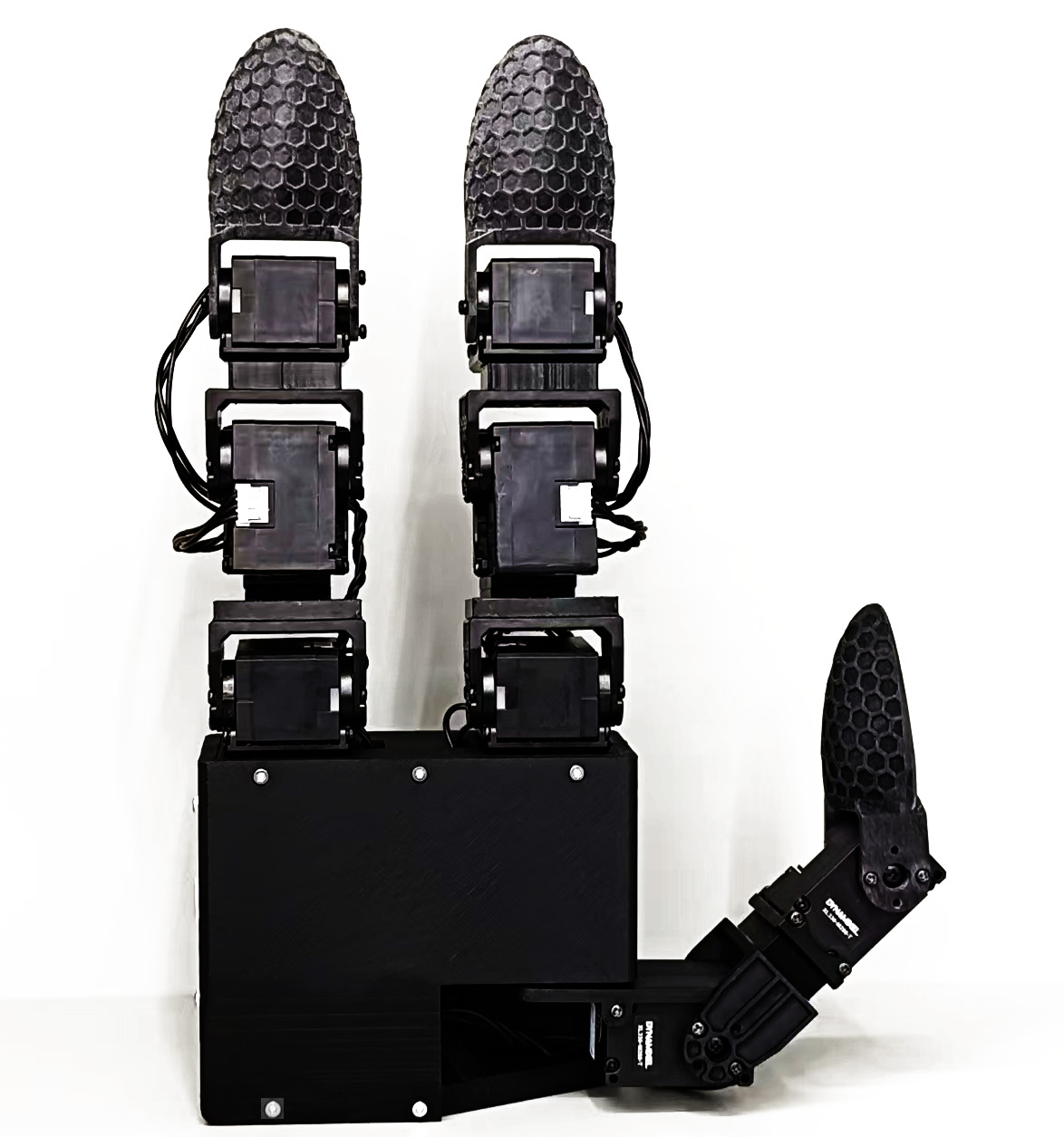}
    \caption{The physical diagram of the GX11 hand.}
    \label{fig:gx11}
\end{figure}

Building upon the Leap Hand~\cite{leaphand} design, We present a redesigned three-finger anthropomorphic fully-driven dexterous hand named GX11. 
The GX11 hand adopts the same direct joint drive method as the Leap Hand, featuring a 3 DoF structure for the thumb while employing identical 4 DoF configurations for both the index and middle fingers, collectively forming a total of 11 DoF. 
This innovative design not only simplifies the mechanical architecture but also achieves significant reductions in overall weight and physical dimensions.
%
The joint configuration and physical diagram of the GX11 hand are presented in Fig.~\ref{fig:gx11_config} and Fig.~\ref{fig:gx11}.
The motion space of the fingertip positions for the three fingers of the GX11 hand is shown in Fig.~\ref{fig:gx11_range}. 

\begin{figure*}[t] 
  \centering
  \begin{minipage}[t]{0.32\textwidth}
    \centering
    \includegraphics[width=\linewidth]{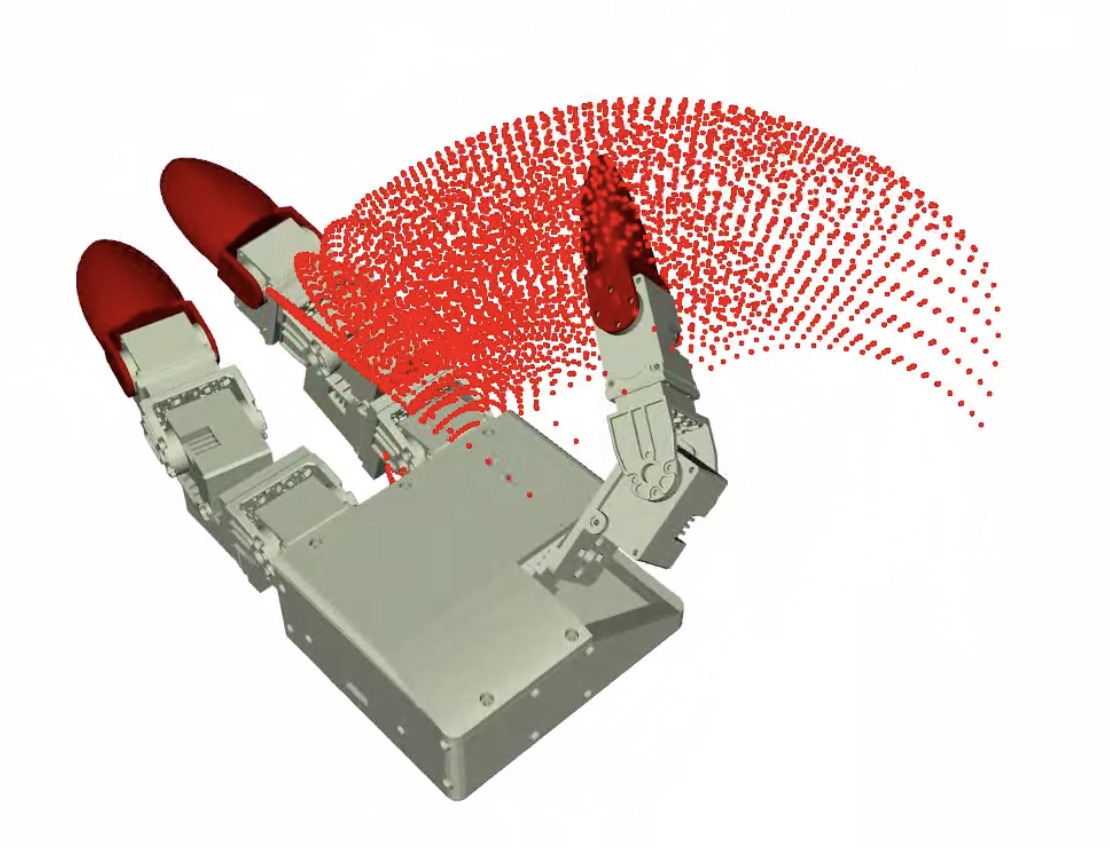}
    \subcaption{Thumb finger}
    
  \end{minipage}
  \hfill
  \begin{minipage}[t]{0.32\textwidth}
    \centering
    \includegraphics[width=\linewidth]{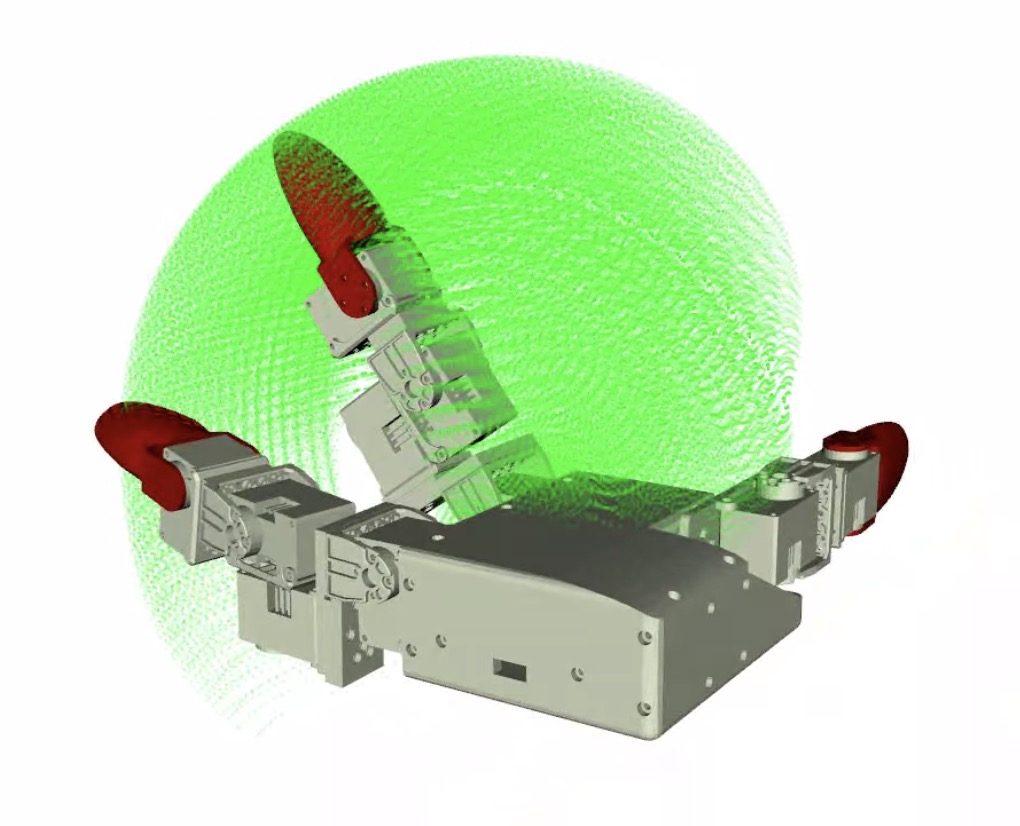}
    \subcaption{Index finger}
    
  \end{minipage}
  \hfill
  \begin{minipage}[t]{0.32\textwidth}
    \centering
    \includegraphics[width=\linewidth]{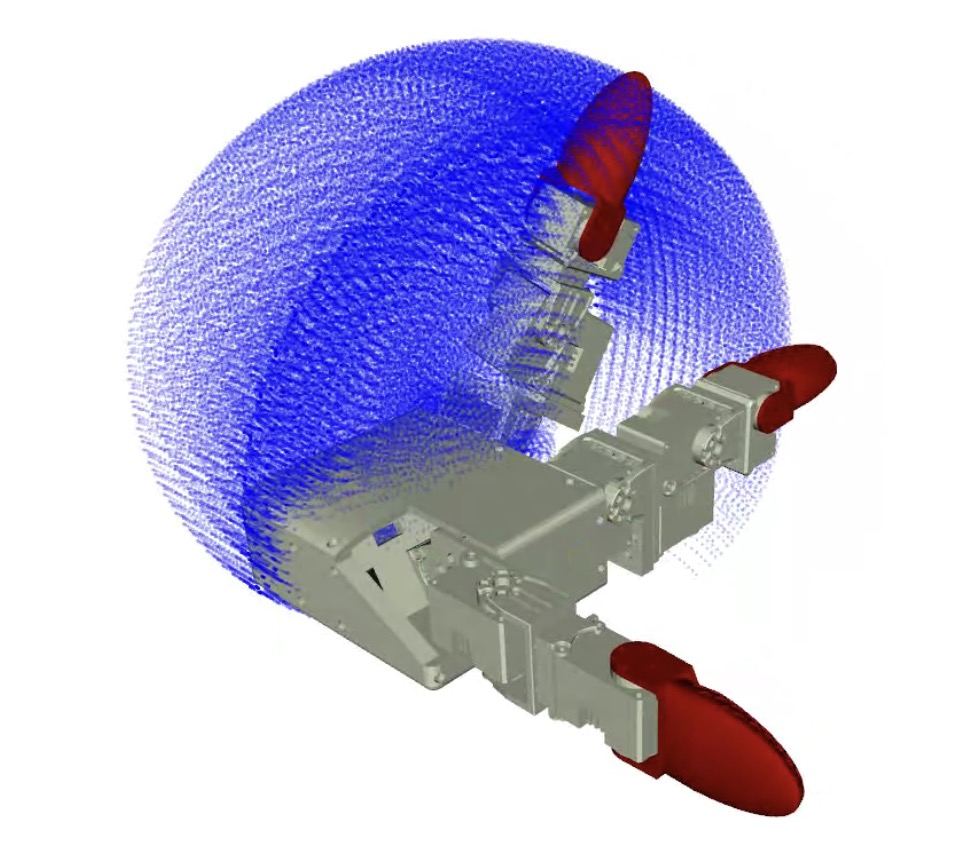}
    \subcaption{Middle finger}
    
  \end{minipage}
  \caption{The motion space of the fingertip positions for the three fingers of the GX11 hand.}
  \label{fig:gx11_range}
\end{figure*}

\begin{figure*}[htbp] 
  \centering
  \begin{minipage}[t]{0.32\textwidth}
    \centering
    \includegraphics[width=\linewidth]{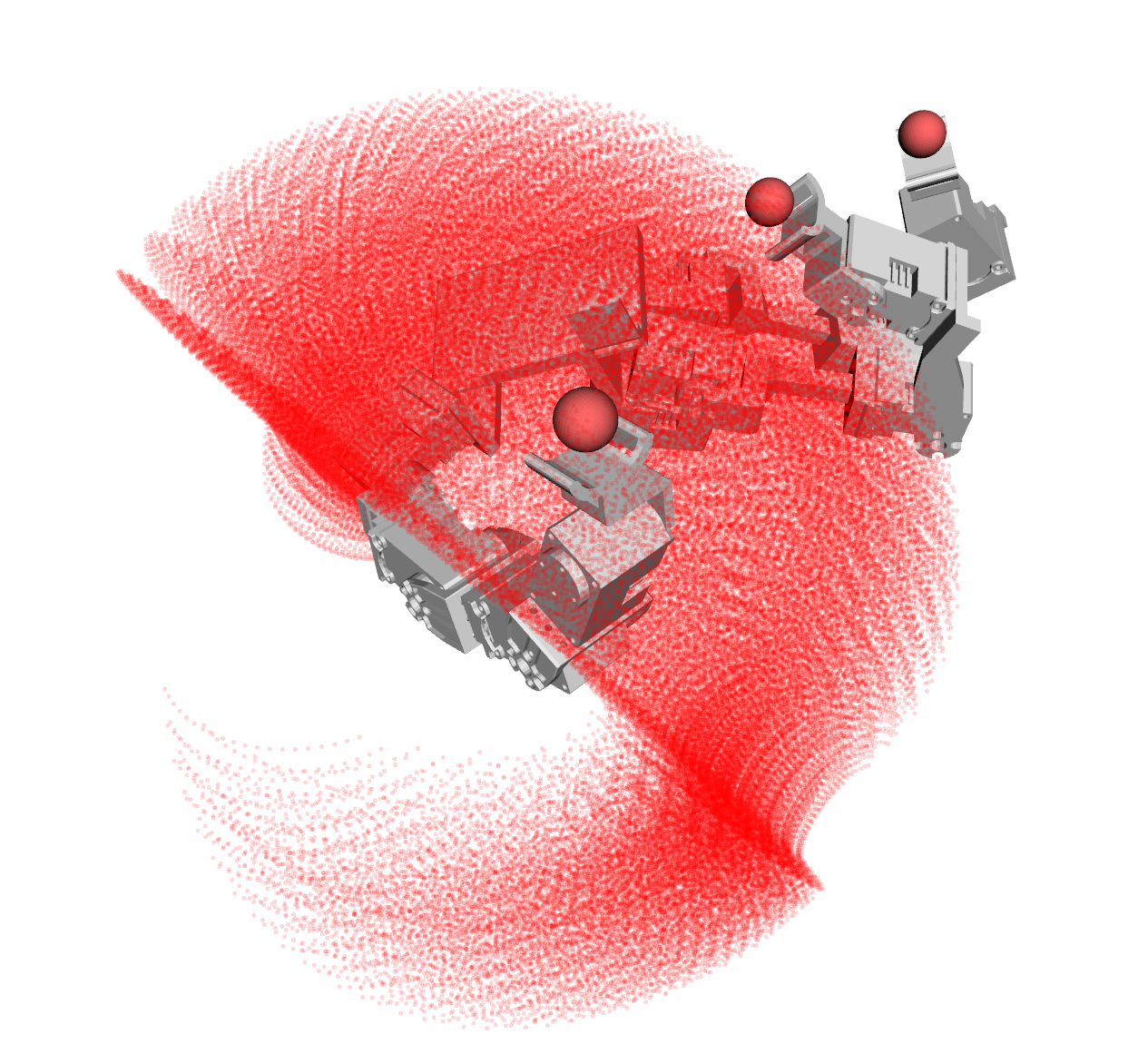}
    \subcaption{Thumb finger}
    
  \end{minipage}
  \hfill
  \begin{minipage}[t]{0.32\textwidth}
    \centering
    \includegraphics[width=\linewidth]{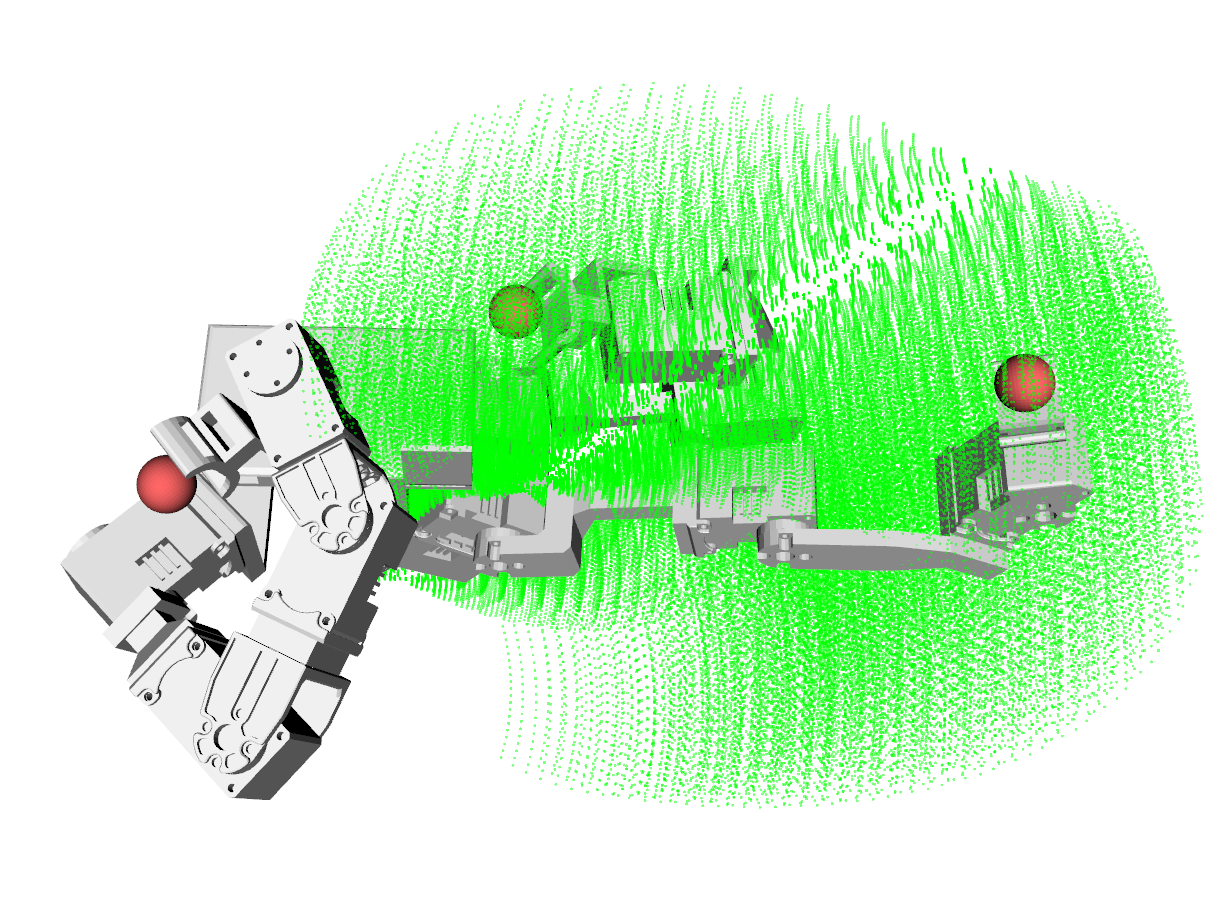}
    \subcaption{Index finger}
    
  \end{minipage}
  \hfill
  \begin{minipage}[t]{0.32\textwidth}
    \centering
    \includegraphics[width=\linewidth]{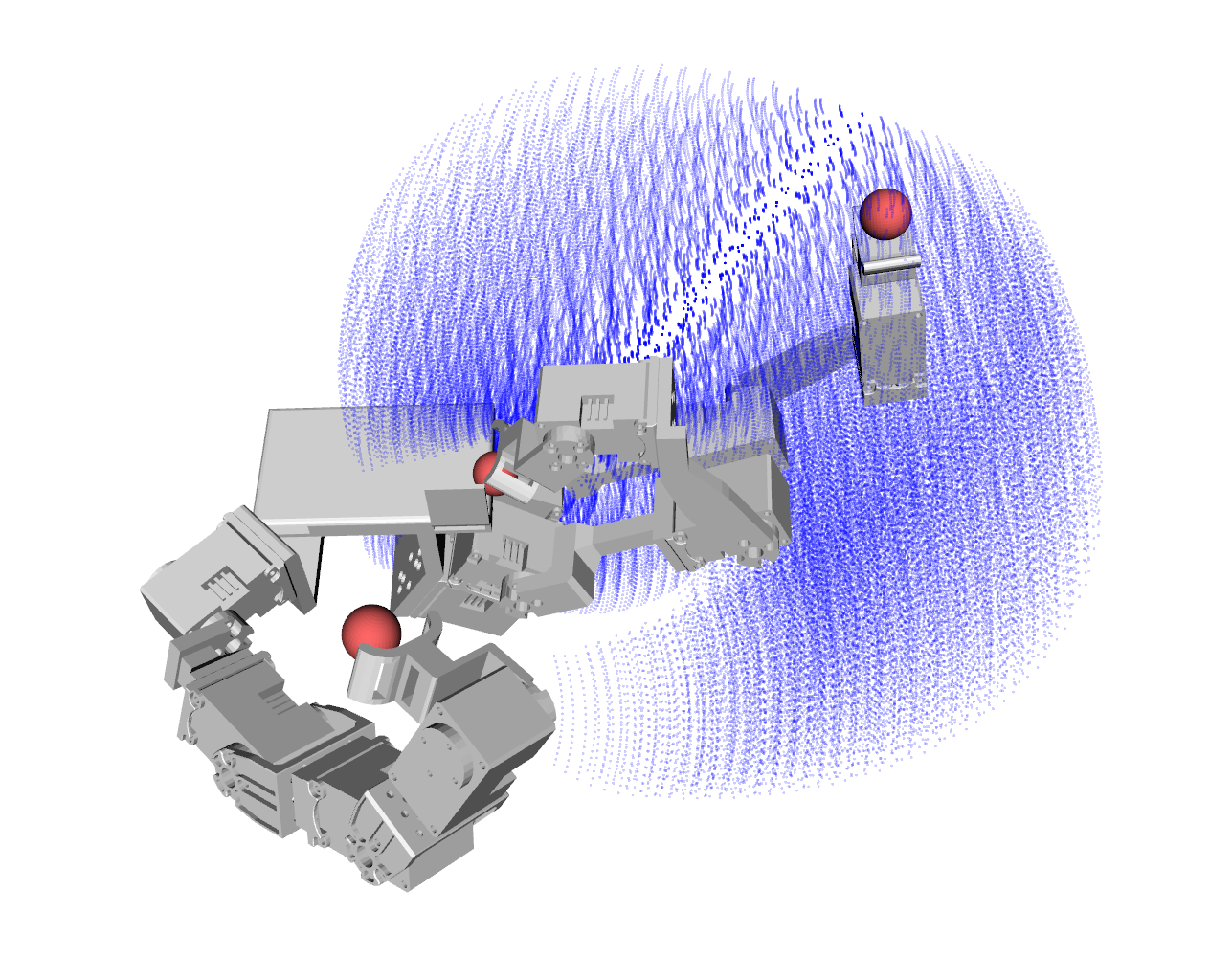}
    \subcaption{Middle finger}
    
  \end{minipage}
  \caption{The motion space of the fingertip positions for the three fingers of the EX12 exoskeleton glove.}
  \label{fig:gx11_range}
\end{figure*}

The GX11's joint actuation employs Dynamixel XL330-M288-T, a high-performance, cost-effective servo actuator. 
These XL330-M288-T units deliver 0.53Nm rated torque, providing sufficient power density for driving all phalangeal joints. 
The integrated 12-bit absolute magnetic encoder (AS5601) ensures high-precision position feedback with 0.088° resolution, guaranteeing motion accuracy across all articulations. 
Through a 288:1 reduction ratio, the actuators achieve enhanced torque amplification while maintaining operational stability and reliability during diverse manipulation tasks.
Operating at 5V DC, the motor system utilizes an ARM Cortex-M0+ microcontroller for real-time control, implementing TTL Multidrop Bus communication protocol that simplifies circuit architecture while improving system scalability and hardware compatibility. 
The actuators support multiple control modalities including current control mode, velocity control mode and position control mode.
This multi-modal flexibility enables adaptive parameter tuning for various operational requirements. With a total mass of merely 18g and compact dimensions (24.5 × 33 × 25 mm), the optimized form factor permits seamless integration into the anthropomorphic joint design without compromising dexterity.

\subsection{Exoskeleton Glove}
\begin{figure}
    \centering
    \includegraphics[width=0.6\linewidth]{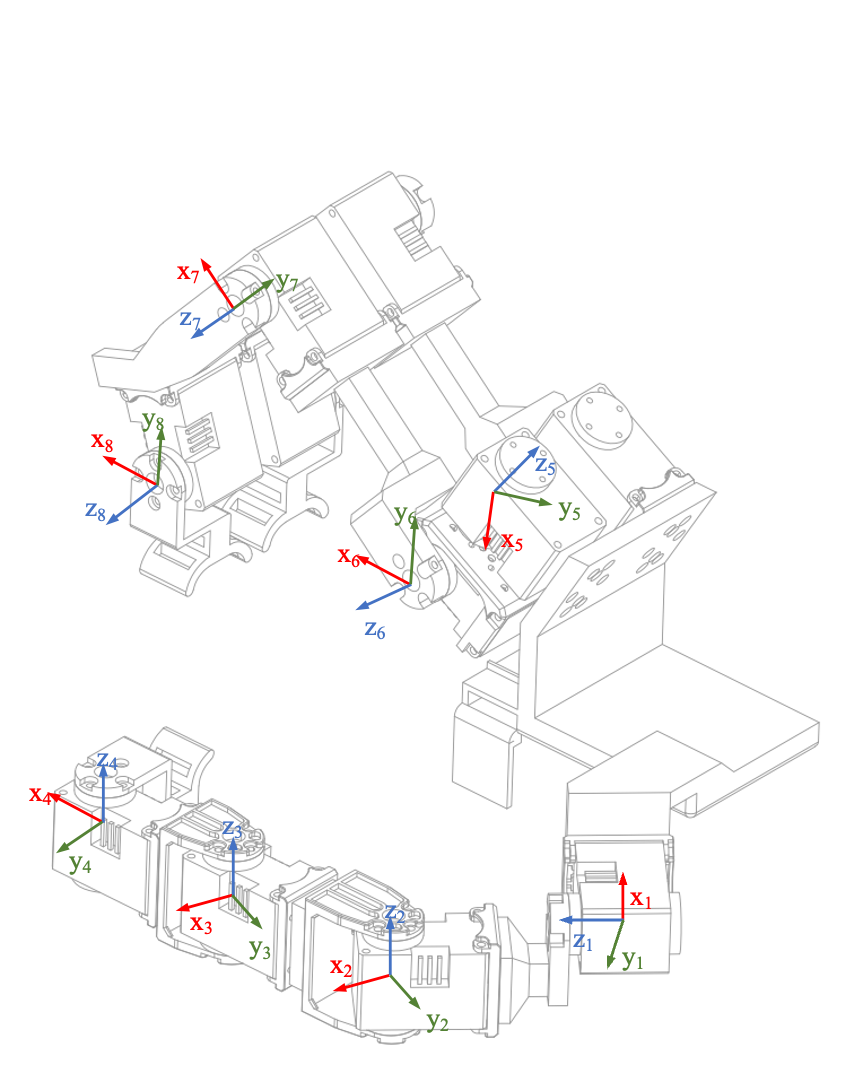}
    \caption{The joint configuration of the EX12 exoskeleton glove.}
    \label{fig:ex12_config}
\end{figure}

\begin{figure}
    \centering
    \includegraphics[width=0.6\linewidth]{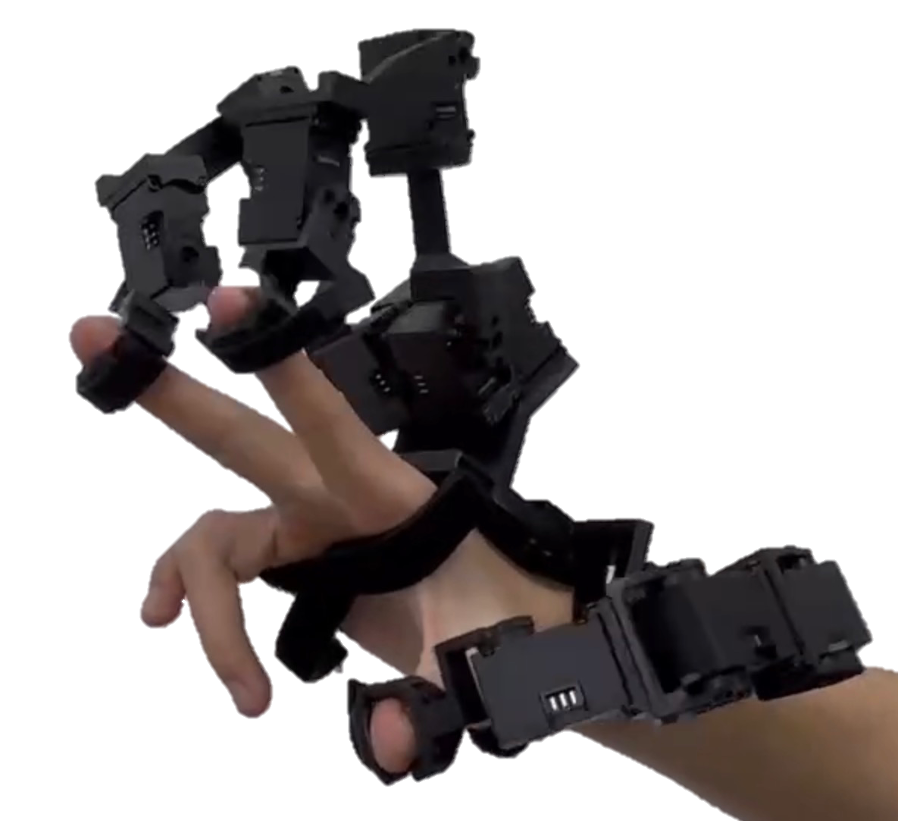}
    \caption{The physical demonstration of the EX12 exoskeleton glove worn on a human hand}
    \label{fig:ex12}
\end{figure}

The EX12 exoskeleton glove adopts Dynamixel XL330-M077-T motor, which has a smaller reduction ratio (77:1) to reduce the friction from the gear box during human finger motion.
%
Considering the high DoF of the human thumb, the EX12 exoskeleton glove incorporates a 4-DoF thumb design to better accommodate the comfort requirements of human teleoperation. 
The other two fingers maintain the same 4-DoF configuration as the GX11 hand.
The joint configuration and physical diagram of the EX12 exoskeleton glove are presented in Fig.~\ref{fig:ex12_config} and Fig.~\ref{fig:ex12}.
The motion space of the fingertip positions for the three fingers of the EX12 exoskeleton glove is shown in Fig.~\ref{fig:gx11_range}.

In the GX11 dexterous hand and EX12 exoskeleton glove, all structural components except the joint motors are fabricated using acrylonitrile butadiene styrene (ABS) material through 3D printing.
ABS demonstrates notable advantages including cost-effectiveness, high mechanical strength, and excellent wear resistance, making it particularly suitable for manufacturing structural components of dexterous hands. 
The adoption of 3D printing structural components exhibit simplified assembly procedures and maintenance accessibility, thereby significantly reducing both overall manufacturing costs and maintenance complexity.


\section{GEX Software}
\subsection{Control SDK}
To achieve unified control of the 11 joint motors in the GX11 dexterous hand and the EX12 Exoskeleton glove, the embedded OpenRB-150 motor control board is employed for motor communication and control. 
The OpenRB-150 represents a high-performance, low-power motor control circuit board specifically designed for multi-motor systems. 
This control board incorporates an ARM Cortex-M0+ master control chip operating at 48 MHz clock frequency, enabling rapid processing of complex control instructions and real-time data. 
Furthermore, the OpenRB-150 supports communication frequencies up to 1M bps, ensuring high-speed and stable communication connections.

We open source a Python control library \texttt{libgex}\footnote{\url{https://github.com/Democratizing-Dexterous/libgex}}, which can communicate with OpenRB-150 motor control board to achieve joint control and joint sensing. 
%
The code snippet of \texttt{libgex}, as shown in Code~\ref{code:handcontrol}.
\texttt{libgex} is implemented in pure Python, featuring high-level abstraction and encapsulation for simplified control. 
It demonstrates cross-platform compatibility with seamless deployment on both Linux and Windows systems.

\subsection{Dexterous Retargeting}
Due to the structural inconsistency between the dexterous hand and the exoskeleton glove, dexterous retargeting is required during teleoperation to accurately map the EX12-worn human hand movements onto the GX11's joint space. 
To address this, we developed a retargeting system based on \texttt{Dex-Retargeting}\footnote{\url{https://github.com/dexsuite/dex-retargeting}}, with the source code being open-sourced\footnote{\url{https://github.com/Democratizing-Dexterous/gex_retargeting_sim}}.

\begin{figure}
    \centering
    \includegraphics[width=0.96\linewidth]{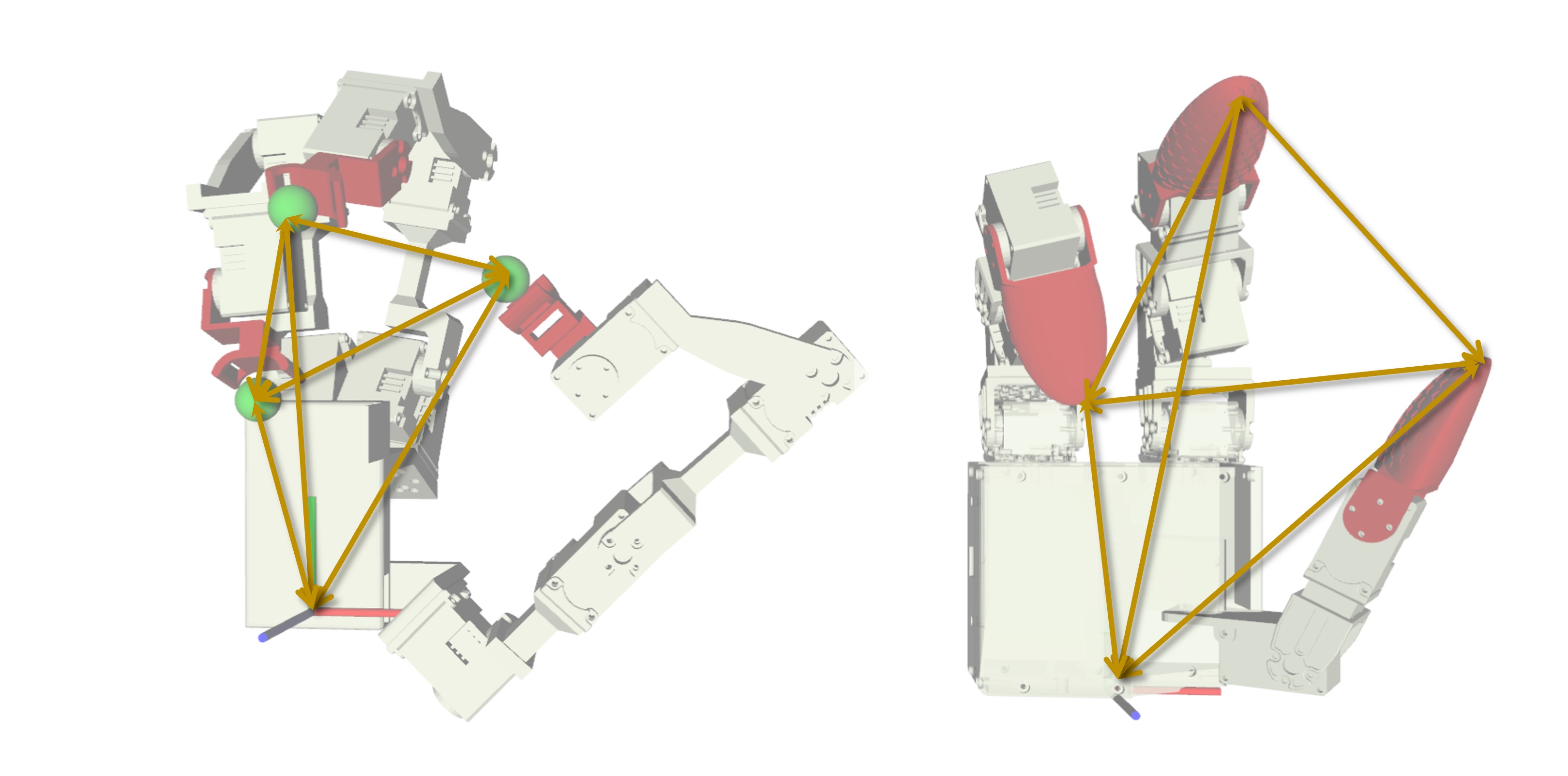}
    \caption{Task space vectors between fingertips and palm for both the teleoperation glove and GX11 pro hand used for retargeting optimization.}
    \label{fig:glove_retarget}
\end{figure}

\begin{figure*}[htbp]
    \centering
    \includegraphics[width=0.97\linewidth]{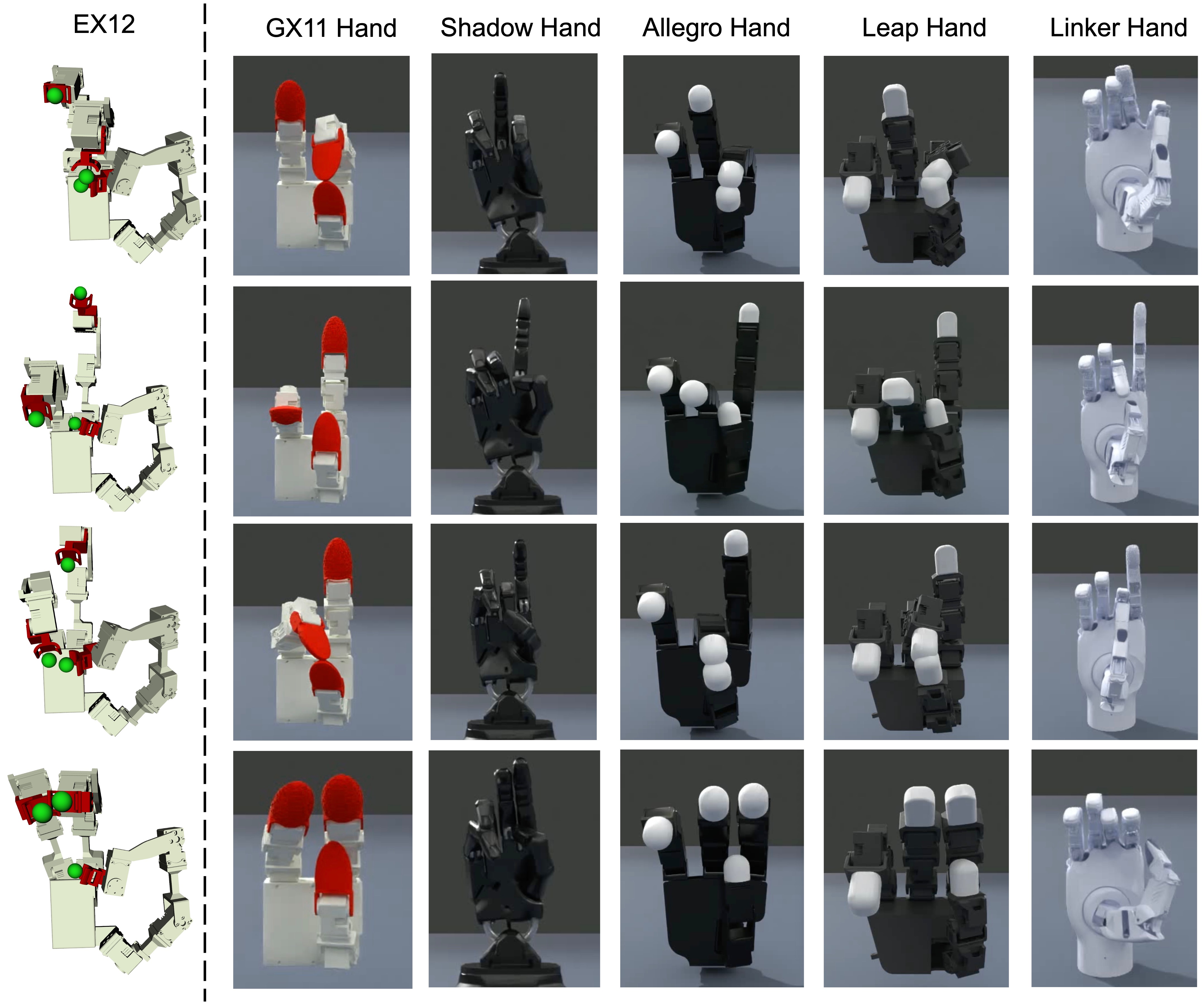}
    \caption{Visualization of retargeting results in Isaac from glove to dexterous hand. 
    The first row shows the glove poses, and the second row shows the retargeted results on the dexterous hand.
    The green spheres in the glove represent the positions of the human fingertips.
    }
    \label{fig:glove_retarget_rst}
\end{figure*}

The retargeting is framed as constrained, with the objective of finding a suitable joint position trajectory $\textbf{q}^{rbt}_t$ that minimizes the following cost function:
\begin{equation}
\begin{aligned}
& \min_{\mathbf{q}^{rbt}_t} \sum_{i=1}^{n_f}  \| F^{rbt}_{i}(\mathbf{q}^{rbt}_t) - \alpha F^{exo}_{i}(\mathbf{q}_t^{exo})  \|^2 + k \| \mathbf{q}_t^{rbt} - \mathbf{q}_{t-1}^{rbt} \|^2 \\
& \text{s.t.} \quad q_l \leq q^{rbt}_t \leq q_u,
\end{aligned}
\end{equation}
where the first summation term enforces spatial pose consistency by minimising the squared deviation between the fingertip pose of the \(i^{\text{th}}\) finger of the dexterous hand, 
\(F^{rbt}_{i}(\mathbf{q}^{rbt}_t)\), and the corresponding fingertip pose of the exoskeleton, \(F^{exo}_{i}(\mathbf{q}^{exo}_t)\); here \(n_f\) denotes the number of force channels and \(\alpha\in\mathbb{R}^{+}\) is a dimension-matching scaling factor.  
The second term \(k\lVert\mathbf{q}^{rbt}_t-\mathbf{q}^{rbt}_{t-1}\rVert^{2}\) penalises inter-step variations in joint positions with a smoothing weight \(k>0\), thereby suppressing trajectory jitter and enhancing executability.  
The bound constraint \(q_l\le q^{rbt}_t\le q_u\) (with \(q_l,q_u\in\mathbb{R}^{n_q}\) denoting joint lower and upper limits, respectively) explicitly guarantees adherence to the robot’s mechanical and safety limits. 
The keypoint placement of the exoskeleton and the dexterous hand is shown in Fig.~\ref{fig:glove_retarget}, where the keypoints are located at the fingertips and at the base of the palm.
The retargeting results of the exoskeleton under different hand poses are illustrated in Fig.~\ref{fig:glove_retarget_rst}. 
As shown in the figure, the exoskeleton motions in various configurations are consistently mapped to the corresponding postures of the dexterous hand, demonstrating the effectiveness of the retargeting scheme.

\begin{figure}
    \centering
    \includegraphics[width=0.9\linewidth]{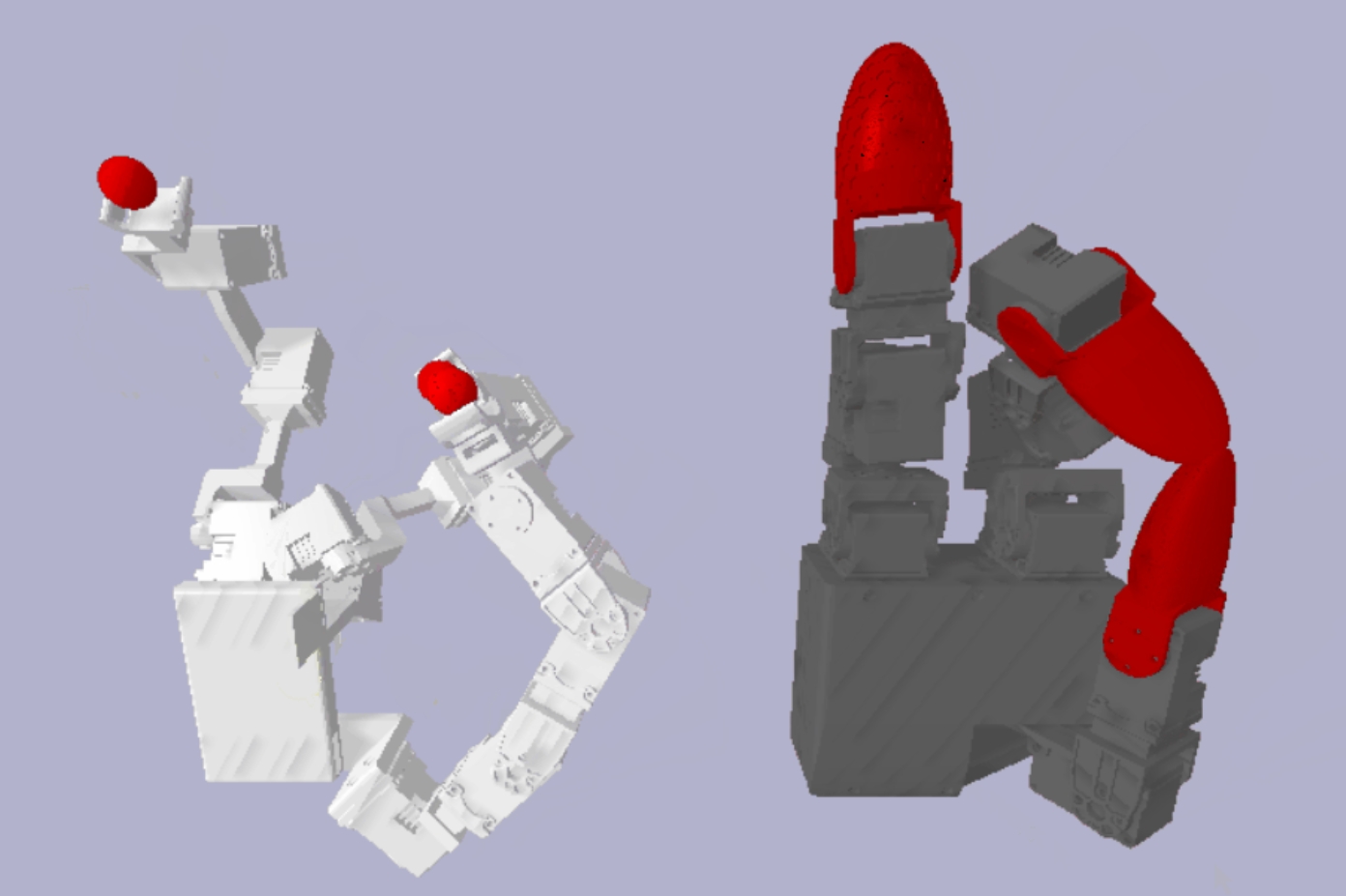}
    \caption{The retargeting results in the PyBullet simulation environment.}
    \label{fig:dex-retargeting}
\end{figure}

%
The retargeting results in the PyBullet simulation environment are illustrated in Fig.~\ref{fig:dex-retargeting}.
%
As evident from the figure, the dexterous retargeting algorithm achieves precise mapping of pinching operations, which is crucial for dexterous teleoperation tasks.

\begin{lstlisting}[language=Python, caption={The python snippets of \texttt{libgex}.}, label=code:handcontrol]
from libgex.libgx11 import Hand
 
hand = Hand(port='/dev/ttyACM0') # COM* for Windows, ttyACM* or ttyUSB* for Linux
hand.connect(goal_pwm=600) # will torque on all the motors

hand.home() # home the hand

hand.motors[0].setj(90) # unit degree

print(hand.getj()) # print all the joints angles

from libgex.libex12 import Glove

glove = Glove('/dev/ttyACM1') # COM* for Windows, ttyACM* or ttyUSB* for Linux, if in Linux, check the port number with `ls /dev/tty*`, then run `sudo chmod 777 /dev/ttyACM*`
glove.connect(init=False) # do not torque on glove yet.

print(glove.fk_finger1()) # get the thumb tip xyz position in base_link frame (bottom of the palm), unit m

\end{lstlisting}

\subsection{Force Feedback}
\begin{figure}
    \centering
    \includegraphics[width=0.9\linewidth]{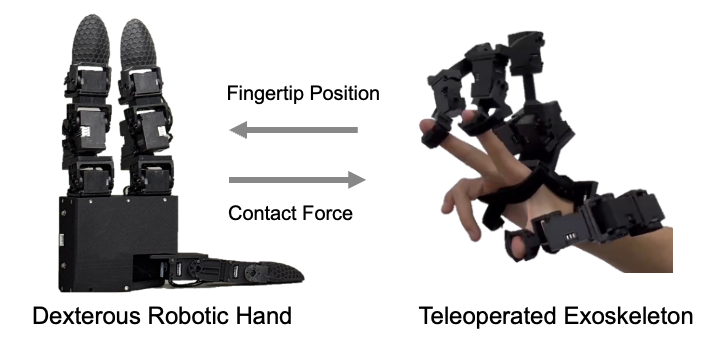}
    \caption{
    The force feedback teleoperation logic of the exoskeleton glove and the dexterous hand.
    }
    \label{fig:interaction}
\end{figure}

Leveraging the current control mode of Dynamixel servo motors, precise joint torque control can be achieved through current regulation. 
When operating in torque control mode, this feature establishes the fundamental basis for realizing bidirectional force feedback during teleoperation procedures.
When the dexterous hand is not in contact with external objects, the joint motors do not output torque and only record joint positions.
The fingertip position of the human hand is estimated based on joint positions and then retargeted to the dexterous hand.
When the dexterous hand detects contact with an external object, the corresponding finger of the exoskeleton glove enters impedance mode to provide force feedback to the human hand.
The contact detection can be done installing tactile sensors or by identifying the dynamics through our previous work~\cite{Tanshen2024}.
The force feedback teleoperation logic is shown in Fig.~\ref{fig:interaction}

\section{Experiments}

To demonstrate the effectiveness of GEX in dexterous manipulation tasks, we conducted an experiment where the GX11 hand was tasked with grasping a paper cup.
This experiment showcases the precise teleoperation capabilities enabled by the combination of the GEX hardware and the GEX software stack, including the \texttt{libgex} library and the dexterous retargeting algorithm. 
%

\begin{figure}
    \centering
    \includegraphics[width=0.9\linewidth]{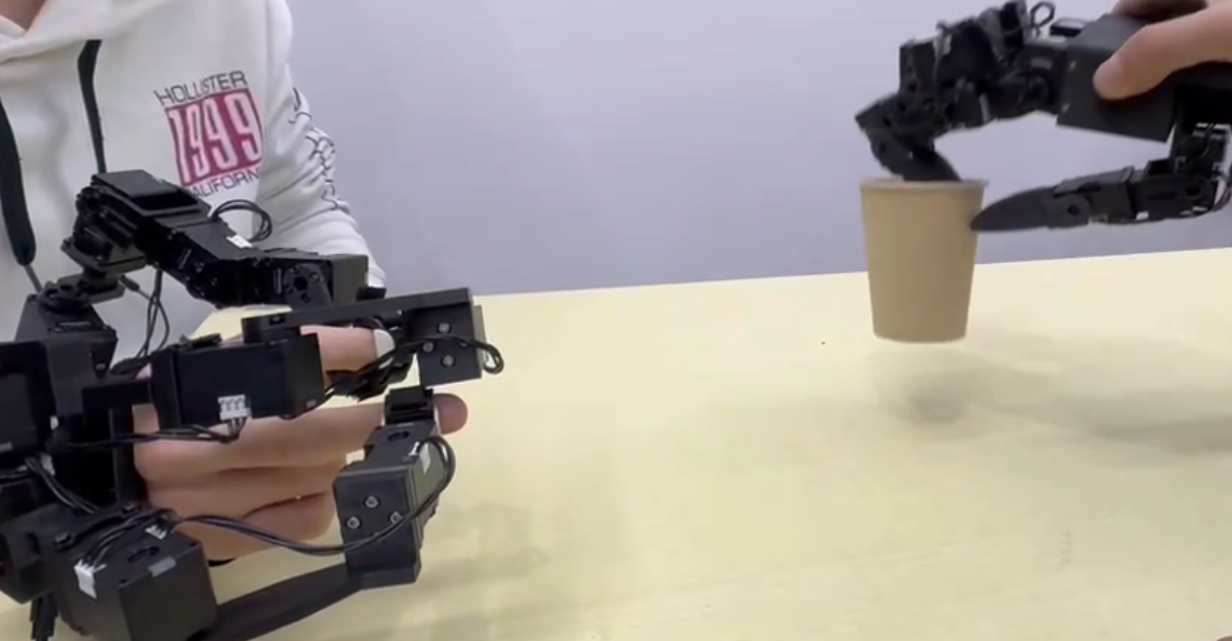}
    \caption{The GX11 hand successfully grasping a paper cup controlled by the EX12 exoskeleton glove.}
    \label{fig:real grasping}
\end{figure}

\begin{figure*}[htbp]
    \centering
    \includegraphics[width=0.7\linewidth]{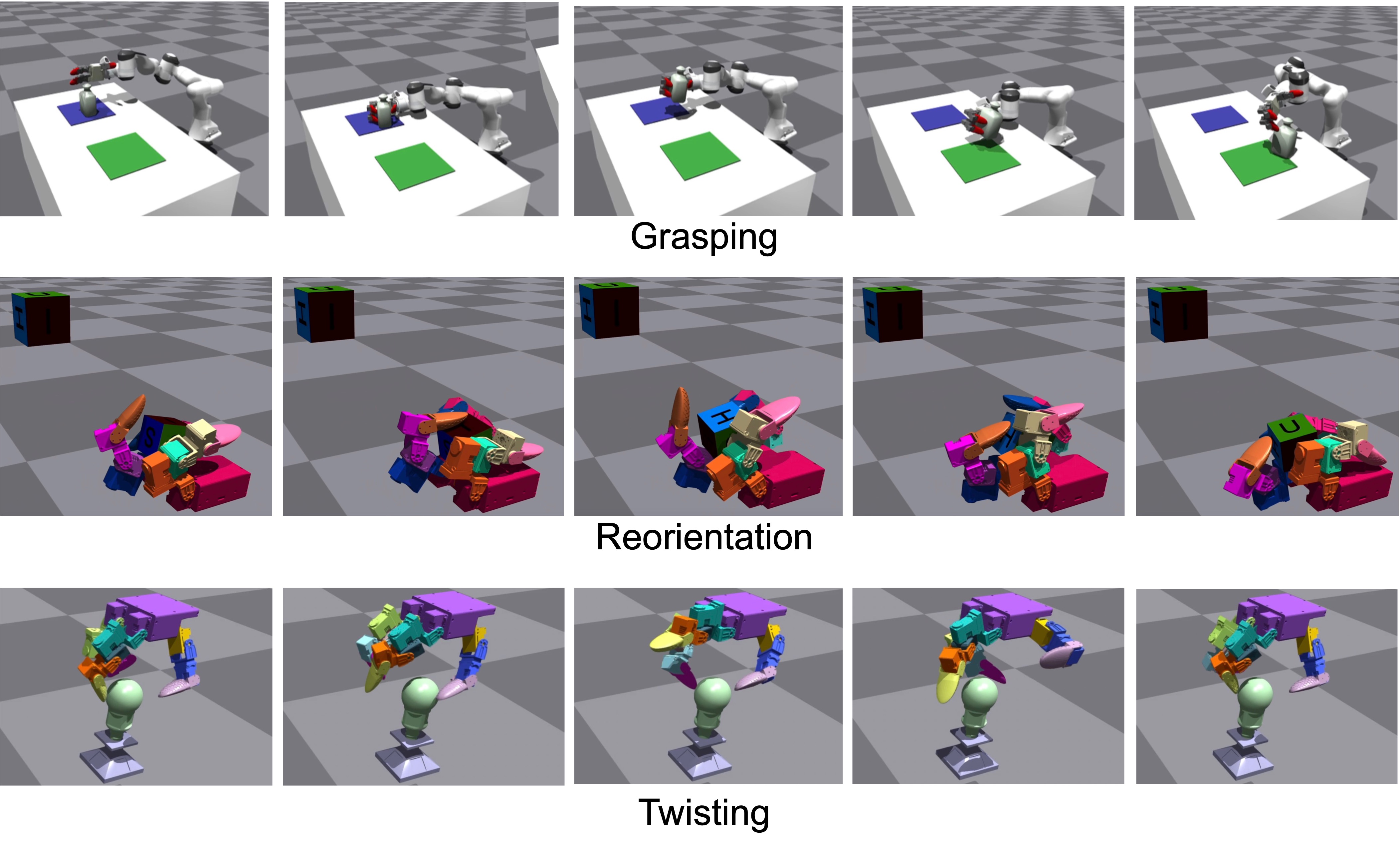}
    \caption{Illustration of the three manipulation tasks using GX11 hand: grasping, reorientation, and twisting. The top row shows table-top grasping, where the robot picks up a target object from the table and moves it to a goal region. The middle row shows object reorientation on the table. The bottom row shows in-hand twisting of an object mounted on a stand. Our dexterous hand is able to successfully accomplish all three tasks within a unified control framework.}
    \label{fig:sim_task}
\end{figure*}

The experiment utilized the complete GEX system, comprising the GX11 hand and the EX12 exoskeleton glove. 
The \texttt{libgex} library was employed to establish communication between the control computer and the OpenRB-150 motor control board, which managed the motor commands for both the hand and the glove. 
The dexterous retargeting algorithm was implemented to map the motion captured by the EX12 exoskeleton glove to the corresponding joint movements of the GX11 hand. This setup enabled real-time teleoperation of the robotic hand based on the user's hand movements.

The experiment concluded successfully with the GX11 hand grasping the paper cup as shown in Fig.~\ref{fig:real grasping}. 
This result underscores the efficacy of the GEX system in enabling precise and intuitive teleoperation for dexterous tasks. 
The integration of the \texttt{libgex} library and the dexterous retargeting algorithm allowed for seamless motion translation from the glove to the hand, highlighting the potential of the GEX platform for applications in robotic manipulation and skill transfer.

\subsection{Manipulation tasks}

As illustrated in Fig.~\ref{fig:sim_task}, we evaluate our method on three representative dexterous manipulation tasks: grasping, reorientation, and twisting. In the grasping task, the robot arm and dexterous hand must approach a target object on the table, establish a stable grasp, and transport the object to a designated goal area. In the reorientation task, the hand interacts with a composite object lying on the tabletop and changes its global pose from a random initial configuration to a desired orientation through a sequence of contact-rich motions. In the twisting task, the hand grasps an object mounted on a stand and performs a controlled rotational motion around the specified axis while maintaining a stable grasp. Across these three tasks, our dexterous hand successfully completes the required behaviors, demonstrating its ability to handle both pick-and-place style operations and fine in-hand manipulation.

\section{Conclusion}
This paper presented GEX, a novel low-cost dexterous manipulation system comprising the GX11 robotic hand and the EX12 exoskeleton glove, establishing a fully-actuated teleoperation framework through high-fidelity kinematic retargeting. 
Our modular 3D-printed design achieves unprecedented cost efficiency while maintaining complete actuation across all 23 DoF, enabling precise bidirectional control and state estimation. 
The system's full-actuation architecture overcomes limitations of conventional tendon-driven or underactuated approaches, providing accurate kinematic modeling and control essential for dexterous manipulation tasks. 
Experimental results demonstrate the system's capability in both grasping and fine manipulation scenarios. 
By bridging the accessibility gap in dexterous manipulation research, GEX offers an open platform for advancing robotic skill learning and embodied AI development. 
All designs and software are open-sourced to facilitate community-driven progress in dexterous robotics research.
Future work will focus on expanding the system's sensory capabilities and developing more advanced control algorithms for complex manipulation tasks. 


%




\ifCLASSOPTIONcaptionsoff
  \newpage
\fi



\bibliographystyle{IEEEtran}

\begin{thebibliography}{9}

\bibitem{akkaya2019solving} 
I. Akkaya, M. Andrychowicz, M. Chociej, M. Litwin, B. McGrew, A. Petron, A. Paino, M. Plappert, G. Powell, R. Ribas et al., ``Solving rubik’s cube with a robot hand,'' \textit{arXiv preprint arXiv:1910.07113}, 2019.

\bibitem{chi2024universal} 
C. Chi, Z. Xu, C. Pan, E. Cousineau, B. Burchfiel, S. Feng, R. Tedrake, and S. Song, ``Universal manipulation interface: In-the-wild robot teaching without in-the-wild robots,'' \textit{arXiv preprint arXiv:2402.10329}, 2024.

\bibitem{shadowhand_inhand} 
O. M. Andrychowicz, B. Baker, M. Chociej, R. Jozefowicz, B. McGrew, J. Pachocki, A. Petron, M. Plappert, G. Powell, A. Ray et al., ``Learning dexterous in-hand manipulation,'' \textit{The International Journal of Robotics Research}, vol. 39, no. 1, pp. 3–20, 2020.

\bibitem{leaphand} 
K. Shaw, A. Agarwal, and D. Pathak, ``Leap hand: Low-cost, efficient, and anthropomorphic hand for robot learning,'' \textit{arXiv preprint arXiv:2309.06440}, 2023.

\bibitem{qin2023anyteleop} 
Y. Qin, W. Yang, B. Huang, K. Van Wyk, H. Su, X. Wang, Y.-W. Chao, and D. Fox, ``Anyteleop: A general vision-based dexterous robot arm-hand teleoperation system,'' \textit{arXiv preprint arXiv:2307.04577}, 2023.

\bibitem{Tanshen2024}
T. Shen, X. Liu, Y. Dong, L. Yang and Y. Yuan, ``Switched Momentum Dynamics Identification for Robot Collision Detection''. IEEE Transactions on Industrial Informatics, 2024.

\end{thebibliography}
\end{document}